\pdfoutput=1
\documentclass[11pt]{article}

\usepackage[preprint]{acl}

\usepackage{times}
\usepackage{latexsym}
\usepackage{balance}
\usepackage[T1]{fontenc}
\usepackage[utf8]{inputenc}

\usepackage{microtype}

\usepackage{inconsolata}

\usepackage{amsmath}
\usepackage{amssymb}
\usepackage{mathtools}
\usepackage{amsthm}

\usepackage{pifont}

\usepackage{graphicx}
\usepackage{adjustbox}
\usepackage{listings}
\usepackage[dvipsnames]{xcolor}
\usepackage{amsmath}
\usepackage{multirow, multicol}
\usepackage{array}
\usepackage{tabularray}
\usepackage{makecell}
\usepackage{bm}
\usepackage{xcolor}
\usepackage[capitalize,noabbrev]{cleveref}
\usepackage{tcolorbox}
\usepackage{longtable}
\usepackage{tabularx}
\usepackage{lipsum}

\theoremstyle{plain}

\theoremstyle{definition}

\theoremstyle{remark}

\newcommand{\myNum}[1]{(\emph{#1})}
\newcommand{\oursys}{$\mathsf{ZeroToD}$}

\newcommand{\llamai}{Llama-3.2}
\newcommand{\flan}{FLAN-T5}
\newcommand{\gpt}{GPT-2}

\newcommand{\apicall}{API Call}
\newcommand{\simpletod}{SimpleTOD}
\newcommand{\autotod}{$\mathsf{Auto}$\mbox{-}$\mathsf{ToD}$}
\newcommand{\zstod}{$\mathsf{ZS}$\mbox{-}$\mathsf{ToD}$}
\newcommand{\soloist}{SOLOIST}
\newcommand{\ood}{out-of-domain}
\newcommand{\bleu}{BLEU-4}

\newcommand{\mycolor}[2]{\leavevmode\textcolor{#1}{#2}}
\newcommand{\cmark}{\ding{51}}%
\newcommand{\xmark}{\ding{55}}%

\newcolumntype{L}{!{\vrule width 1.1pt}c}
\newcolumntype{R}{c!{\vrule width 1.1pt}}
\newcommand{\thickhline}{\noalign{\hrule height 1.1pt}}
\newcommand{\failed}{\ding{55} (Failed to generate a valid response)}

\title{Evaluating and Enhancing Out-of-Domain Generalization of Task-Oriented Dialog Systems for Task Completion without Turn-level Dialog Annotations}

\author{Adib Mosharrof \quad Moghis Fereidouni \quad A.B. Siddique\\
  University of Kentucky \\
  Lexington, KY, USA \\
  \small\texttt{\{adib.mosharrof, moghis.fereidouni, ab.siddique\}@uky.edu}
}

\begin{document}
\maketitle
\begin{abstract}

Traditional task-oriented dialog (ToD) systems rely heavily on labor-intensive turn-level annotations, such as dialogue states and policy labels, for training. 
This work explores whether large language models (LLMs) can be fine-tuned solely on natural language dialogs to perform ToD tasks, without requiring such annotations. We evaluate their ability to generalize to unseen domains and compare their performance with models trained on fully annotated data.
Through extensive experiments with three open-source LLMs of varying sizes and two diverse ToD datasets, we find that models fine-tuned without turn-level annotations generate coherent and contextually appropriate responses.
However, their task completion performance -- measured by accurate execution of {\apicall}s -- remains suboptimal, with the best models achieving only around 53\% success in unseen domains.
To improve task completion, we propose {\oursys}, a framework that incorporates a schema augmentation mechanism to enhance {\apicall} accuracy and overall task completion rates, particularly in out-of-domain settings. 
We also compare {\oursys} with fine-tuning-free alternatives, such as prompting off-the-shelf LLMs, and find that our framework enables smaller, fine-tuned models that outperform large-scale proprietary LLMs in task completion. 
Additionally, a human study evaluating informativeness, fluency, and task completion confirms our empirical findings. 
These findings suggest the feasibility of developing cost-effective, scalable, and zero-shot generalizable ToD systems for real-world applications.

\end{abstract}

\section{Introduction}

% \begin{figure}[!t]
%     \centering
%     \includegraphics[width=0.97\linewidth]{assets/NL_ToD.pdf}
%     \vspace{-6pt}
%     \caption{
% This work leverages natural language conversational data alone, eliminating the need for extensive manual annotations required by existing works.
% It achieves generalization to new, unseen domains through the use of domain schemas.}
%     \label{fig:approach}
%     \vspace{-14pt}
% \end{figure}

Task-oriented dialog (ToD) systems~\cite{zhang2020task} enable users to accomplish diverse tasks through natural language interactions. These systems power virtual assistants, customer service chatbots, and various other applications such as making reservations or scheduling appointments~\cite{Williams2016TheDS, Zhang2019TaskOrientedDS}. 
To be effective, ToD systems must not only engage in user interactions to collect and provide task-specific information but also interface with external systems to accurately complete user tasks.
%To be effective, ToD systems must engage with users, understand task-specific intents, and take appropriate actions to fulfill user requests.

Traditionally, the development of ToD systems has relied heavily on turn-level manually annotated conversational data, where natural language turns are labeled with dialog states and policy actions~\cite{Zhang2020RecentAA}. However, this \emph{reliance on turn-level annotated data limits the scalability of ToD systems}, as it prevents them from fully leveraging the vast amounts of readily available unannotated task-oriented conversational data. Furthermore, the annotation process is labor-intensive, expensive, and prone to inconsistencies and errors~\cite{eric2020multiwoz,zang2020multiwoz,han2021multiwoz,budzianowski2019challenges}.

Recent advancements in natural language processing, particularly the emergence of pre-trained large language models (LLMs)~\cite{vaswani2017attention,devlin2019bert,radford2019language}, offer new opportunities to address these scalability challenges. LLMs have demonstrated remarkable capabilities in diverse language tasks, from understanding context to generating coherent responses. 
While pre-trained models (e.g., GPT-2) have been employed to develop ToD systems~\cite{hosseini2020simple,Yang2020UBARTF,Mosharrof2023ZeroShotGE,budzianowski2018towards}, their potential to train ToD systems without turn-level annotations remains largely unexplored, as does their \emph{ability to generalize effectively to unseen domains}. %Evaluating and improving {\ood} generalization of ToD systems is critical for their real-world deployments.
%, as predefined domains and intents cannot be guaranteed in real-world settings.

%In addition to engaging with users through natural language interactions -- such as requesting task-specific information or providing updates -- a critical aspect of ToD systems is their ability to interact with external systems (e.g., databases) to ensure successful task completion. 
Beyond natural language interactions -- such as requesting task-specific information or providing updates -- ToD systems must also interact with external systems (e.g., databases) to ensure successful task completion.
This often requires retrieving information or executing actions, such as making a reservation via \emph{an \apicall}. 
While ToD systems described in the literature could, in theory, be trained to make such {\apicall}s, this capability \emph{is rarely evaluated in practice}. The lack of rigorous evaluation in this area leaves a significant gap in understanding the readiness of current ToD systems for real-world deployments.

% v1 intro last para

% To overcome the aforementioned challenges, in this paper we propose a TOD approach that enables task completion without requiring additional turn-level annotations, reducing the need for manual data labeling while maintaining strong performance. We employ a multi-task instruction fine-tuning, where the model jointly learns to generate system responses and {\apicall}s, enabling models to learn task completion in a more unified manner. To address the key challenge of poor generalization to unseen domains, we introduce a schema augmentation mechanism by enriching the training data with diverse schema variations, that enhances the ability of models to adapt to new domains. Additionally, we explore whether fine-tuning is necessary for TOD systems by comparing fine-tuned models against fine-tuning-free approaches.

% v2 intro last para

% In this paper, we investigate three fundamental questions in TOD systems: 
%Motivated by the need to evaluate and improve {\ood} generalization of ToD systems, in this work, we investigate three fundamental research questions:

Motivated by the need to evaluate and enhance the {\ood} generalization of ToD systems, this work investigates three research questions:

\noindent
\textbf{RQ1:} Can pre-trained LLMs be adapted into effective ToD systems without turn-level annotated data (e.g., annotated dialog states)?
%ToD systems function effectively ?
%Can ToD systems function effectively without turn-level annotated data?

\noindent
\textbf{RQ2:} How can we improve the {\ood} generalization of ToD systems for task completion?

\noindent
\textbf{RQ3:} How does the {\ood} generalization of fine-tuned ToD systems compare to that of large-scale, proprietary LLMs?
%Is fine-tuning still necessary for ToD systems?

To address RQ1, we frame ToD as a multi-task instruction fine-tuning problem, where the model learns to generate both natural language responses and {\apicall}s by conditioning on the dialog history and domain schema. 
To enhance task completion performance, we introduce a schema augmentation mechanism that enriches training data with diverse schema variations, significantly improving robustness in unseen domains (RQ2). Finally, to investigate RQ3, we compare fine-tuned ToD systems against fine-tuning-free approaches that rely on large-scale, proprietary LLMs, which are often costly and less controllable.

We conduct extensive experiments on two benchmark ToD datasets -- SGD~\cite{Rastogi2019TowardsSM} and KETOD~\cite{Chen2022KETODKT} -- using three open-source models: {\gpt}\cite{radford2019language}, {\llamai}, and {\flan}~\cite{Chung2022ScalingIL}. To provide a comprehensive evaluation, we introduce multiple metrics to assess {\apicall} generation, including method name accuracy, parameter correctness, and complete {\apicall} accuracy. For response generation, we use BERTScore~\cite{Zhang2019BERTScoreET} to better capture the semantic similarity between system outputs and ground truth responses. Additionally, we conduct human studies and qualitative analyses on a subset of both datasets to complement automatic evaluations.

Our empirical results provide clear answers to the research questions posed in this study. For RQ1, we compare our approach against state-of-the-art (SOTA) methods that rely on annotated data and find that ToD systems can function effectively without manual annotations by leveraging multi-task instruction fine-tuning. 
On the complete API accuracy metric, our best model improves by an average of 62.9\% across both datasets compared to the strongest baseline SOTA model trained with turn-level annotated data.
For RQ2, we evaluate the impact of schema augmentation by comparing models trained with and without this mechanism. Our results show that augmentation significantly enhances {\ood} generalization, improving complete API accuracy on unseen domains by 17.05\% for {\flan} and 35.6\% for {\llamai} compared to their non-augmented counterparts.

For RQ3, we compare {\oursys} against fine-tuning-free alternatives in unseen domains and confirm that fine-tuning is advantageous for learning when to make {\apicall}s and maintaining strong {\ood} performance in complex, multi-turn task completion scenarios. 
On complete API accuracy for unseen domains, {\flan} achieves an average improvement of 30.45\% over the best SOTA approach built with the large-scale GPT-4o model.
Furthermore, human study results evaluating informativeness, fluency, and task completion closely align with automatic metrics, confirming our empirical findings. 
\vspace{-8pt}
\section{Related Work}
\vspace{-7pt}

\textbf{Pipeline Approaches.}
ToD systems have traditionally been designed as pipeline systems, where separate components for Natural Language Understanding (NLU), Dialog State Tracking (DST), Dialog Policy, and Natural Language Generation (NLG) are used to handle specific parts of the dialog processing~\cite{ren2018towards, lee2013structured,peng2018deep,le2021predictable,wen2015semantically,peng2020few,chen2019semantically,budzianowski2018towards,Mosharrof2023TowardOS}. However, this approach has drawbacks like error propagation, where errors made in early stages adversely effect modules later on in the pipeline. 

\noindent
\textbf{End-to-End Approaches.}
Recent works have shifted towards E2E learning methods, where the ToD task is formulated as a conditional generation, where the model generates responses based on the entire dialog history and other relevant annotations (e.g., DST)~\cite{hosseini2020simple,Lin2021LeveragingSD,Bang2023TaskOptimizedAF,Zhang2023EnhancingPO,Ham2020EndtoEndNP,Chung2023InstructTODSLL,Yang2020UBARTF,Sun2022BORTBA,Imrattanatrai2023EndtoEndTD,Sun2022MarsSC,Zhao2022AnyTODAP,Peng2021SoloistBT,Mosharrof2023ZeroShotGE,Siddique2022PersonalizingTD}. 
% For example, T5DST \cite{Lin2021LeveragingSD} was introduced a slot description enhanced generative approach for zero-shot ToD. 
% \citet{Zhang2023EnhancingPO} developed FiD-ToD that employs a caching mechanism for the dialog annotations, which can be extracted using a retrieval module.
A major drawback of these approaches is the dependency on manually annotated data, thus limiting the usage of the wealth of available data. Additionally, most of these approaches assume that {\apicall} results are included in the annotated data, thus limiting their ability to evaluate task completion.

\noindent
\textbf{Prompting Approaches.}
Another recent research direction in ToD systems is in-context learning, where pre-trained LLMs are adapted to specific domains based on contextual examples without requiring fine-tuning~\cite{Labruna2023UnravelingCA,Hudevcek2023AreLL,Dingliwal2021FewSD, Madotto2020LanguageMA, Li2022ControllableDS, Madotto2021FewShotBP,Xu2024RethinkingTD}.
Even though these approaches show promise on generic domains, they fail on complex domains, and have specialized structure or requirements. 

% Most works in all the categories require turn-level annotated data to train TOD models, which can be a significant limitation to scale in real world where such data may be scarce or costly.
% Moreover, most systems assume that knowledge from external sources will be provided in the dialog context.
% \emph{Our work, in contrast, focuses on training TOD models using only readily available natural language interactions, while leveraging domain schema for {\ood} generalization.}

% \noindent
% \textbf{Graph Based Approaches.}
% Additionally, some works used graph-based methods to model the flow of dialog, e.g., by representing DST and policy decisions as graphs~\cite{Sohn2023TODFlowMT, Tuan2022TowardsLI, He2023KRPDSAK}. 
% For instance, \citet{He2023KRPDSAK} introduced KRP-DS, a system that incorporates a knowledge graph as external information through a dedicated knowledge module, enabling context-aware path reasoning.
% This graph assists in knowledge prediction, which is used to enhance response generation. 
% \cite{Sohn2023TODFlowMT} introduced the TOD-Flow graph, where a graph is created from the dialog data and annotations, which uncovers the underlying task structure. 
% Unfortunately, these approaches introduce the overhead of building additional modules which makes the process more complex and hard to generalize.
\vspace{-5pt}
\section{Methodology}
\vspace{-4pt}
% \noindent
% \textbf{Problem Formulation.}
\subsection{Problem Formulation}
% \vspace{-2pt}
We formulate ToD task completion as a conditional sequence generation problem, where the system generates natural language responses or {\apicall}s using the dialog history and related domain schemas. 
We leverage domain schema to facilitate out-of-domain generalization in ToD systems.

We formalize the schema for a given domain \(d_x \in D\) by specifying a set of user intents \(\mathcal{I}_{d_x}\).
For example, in the \texttt{Restaurants} domain, one such intent might be \texttt{ReserveRestaurant}.
Each intent \(i_d \in \mathcal{I}_{d_x}\) is then associated with a set of slots \(\mathcal{S}_{i_d}\).
For instance, \texttt{party size} and \texttt{reservation time} might be slots for the \texttt{ReserveRestaurant} intent. 
Each slot \(s_{i} \in \mathcal{S}_{i_d}\) is characterized by a tuple \(\bigl(\text{name}(s),\,\text{is\_required}(s)\bigr)\), indicating the name of slot (e.g., \texttt{reservation date}) and whether it is mandatory to fulfill a desired intent.
% We formalize the schema for a given domain \(d \in D\) by specifying a set of user intents \(\mathcal{I}_d\).
% For example, in the ``Restaurants'' domain, one such intent might be ``Reserve Restaurant''.
% Each intent \(i_d \in \mathcal{I}_d\) is then associated with a set of slots \(\mathcal{S}_{i_d}\).
% For instance, ``party size'' and ``reservation time'' might be slots for the ``Reserve Restaurant'' intent. 
% Each slot \(s \in \mathcal{S}_{i_d}\) is characterized by a tuple \(\bigl(\text{name}(s),\,\text{is\_required}(s)\bigr)\), indicating the name of slot (e.g., ``reservation date'') and whether it is mandatory to fulfill a desired intent.
We denote the entire domain schema for domain \(d_x\) as:
\(
\Gamma_{d_x} 
\;=\; 
\Bigl(d_x,\; \mathcal{I}_d,\;\{\,\mathcal{S}_{i_d}\,\mid\,i_d\in\mathcal{I}_{d_x}\}\Bigr).
\)

A dialog session $\mathcal{T}_i$ of up to $T$ turns is defined as a sequence of user and system utterances:
\(
   \mathcal{T}_i = \bigl((u_1,r_1),\,(u_2,r_2),\,\dots,\,(u_T,r_T)\bigr),
\)
where $u_t$ is the user utterance and $r_t$ is the system response at turn $t$. 
We denote the dialog history at turn $t$ by
\(
   H_t = \{\,(u_1,r_1),\,(u_2,r_2),\dots,(u_{t-1},r_{t-1}),\,u_t\},
\)
which encapsulates all user-system exchanges up to and including the current user utterance \(u_t\). 
Since a single dialog may reference multiple domains, if \(\mathcal{T}_i\) spans \(m\) domains, we write \(\mathcal{T}_i \sim \{\,d_1,\,d_2,\,\dots,\,d_m\} \;\subseteq\; D.
\)
% In our formulation, the system generates its response \(r_t\), at turn \(t\), by conditioning on the relevant domain schemas \(\{\Gamma_{d_j}\}_{j=1}^m\) and the current dialog history \(H_t\).

% \noindent
% \textbf{Schema Augmentation.}
\vspace{-3pt}
\subsection{Schema Augmentation}
Beyond the original set of domain schemas, we create semantic variations of each domain’s intents and slots. Specifically, for each domain \(d_x \in D\), we we define its \(k\)-th schema variant as:
\(
\Gamma_{\tilde{d}^k_x} 
\;=\;
\Bigl(\, \tilde{d}^k_x,
  \tilde{\mathcal{I}}_{d^k_x},\;
  \bigl\{\,
    \tilde{\mathcal{S}}_{i_d}^k
    \;\mid\;
    \tilde{i}_d^k \in \tilde{I}_{d^k_x}
  \bigr\}
\Bigr),
\)
where $\tilde{I}_{d^k_x}$ is the renamed set of intents, and $\tilde{S}_{i_d}^k$ represents the renamed slots for each intent $\tilde{i}_d^k$. 
% For example, in the ``Restaurants'' domain, the original intent ``Reserve Restaurant'' might be changed to ``Reserve Table'', and the slot ``party size'' might become ``number of people''.
% To integrate these augmented schemas into the dialogs, we systematically replace schema references in existing dialogs with their counterparts from \(\Gamma_{\tilde{d}^k} \).
% Concretely, for each dialog $\mathcal{T}_i$ associated with domain \(d\), we construct an augmented dialog $\tilde{\mathcal{T}}_i^k$ by substituting all intents and slots with those from \(\Gamma_{\tilde{d}^k} \). 
% This procedure preserves the underlying dialog flow but exposes {\oursys} to multiple schema variations, ultimately improving its ability to generalize to out-of-domain task scenarios.

For example, in the \texttt{Restaurants} domain, the original intent \texttt{ReserveRestaurant} might be changed to \texttt{ReserveTable}, and the slot \texttt{party size} might become \texttt{number of people}.
To integrate these augmented schemas into the dialogs, we systematically replace schema references in existing dialogs with their counterparts from \(\Gamma_{\tilde{d}^k_x} \).
Concretely, for each dialog $\mathcal{T}_i$ associated with domain \(d_x\), we construct an augmented dialog $\tilde{\mathcal{T}}_i^k$ by substituting all intents and slots with those from \(\Gamma_{\tilde{d}^k_x} \). 
This procedure preserves the underlying dialog flow but exposes {\oursys} to multiple schema variations, ultimately improving its ability to generalize to out-of-domain task scenarios.
\vspace{-3pt}
\subsection{Multi-task Instruction Fine-tuning}
A ToD system must handle diverse interactions, including general conversation, requesting task-specific information, providing details, and making {\apicall}s for task completion. Broadly, the system generates two types of outputs: \myNum{i} natural language responses, and \myNum{ii} {\apicall}s, which include a method name, parameters, and corresponding values. We employ multi-task instruction fine-tuning that trains the model to autonomously decide between generating an {\apicall} or a user response, without introducing special tokens.

Formally, an autoregressive language model (e.g., GPT-2~\cite{radford2019language}) generates text by predicting the next token given the preceding context. For a given sequence of tokens \( (x_1, x_2, \ldots, x_{t-1}) \), the probability distribution for the next token \( x_t \) is computed as:  \(
p(x_t \mid x_{1:t-1}; \theta) = f_{\theta}(x_{1:t-1}),
\)
where \( f_{\theta} \) represents the model parameterized by \( \theta \) and outputs a probability distribution over the vocabulary \( \mathcal{V} \). The next token \( x_t \) is then sampled from this distribution.  
This formulation extends naturally to response generation in ToD systems, where the system response \( r_t \) at turn \( t \) is generated recursively until an end-of-sequence token (\texttt{<eos>}) is produced:  \(
    r_t \sim p(r_t \mid H_t; \theta),
\)
where \( H_t \) denotes the dialog history up to turn \( t \).

To improve out-of-domain generalization, {\oursys} introduces an additional conditioning variable, the domain schema \( \Gamma_{d_x} \) for each domain \( d_x \) and an instruction prompt \( P \). 
The instructions encourage the model to comprehend schema representations to better generalize across unseen domains and dialog contexts. 
Extending the above formulation to multi-task instruction fine-tuning for multi-turn dialogs of length \( T \), where each dialog may span multiple domains \( \{d_1, d_2, \dots, d_m\} \subseteq D \), we optimize the following objective:  
\( - \sum_{t=1}^{T} \log p(r_t \mid P, \{\Gamma_{d_j}\}_{j=1}^{m}, H_t,; \theta).
\)
Since LLMs operate under a finite context length, the dialog history \( H_t \) consists of only the most recent \( k \) turns, where \( k \leq t \).

\vspace{-5pt}
\subsection{Training Details}
\vspace{-3pt}
The dialog history and domain schema are passed through a structured template to form the inputs to the model. 
The template is detailed in Figure~\ref{fig:finetuning_template} in Appendix~\ref{sec:templates}.
Training begins with 500 warm-up steps and early stopping on the evaluation loss with a patience value of 3.
We used the AdamW~\cite{Loshchilov2017DecoupledWD}  optimizer with weight decay and a learning rate of 0.001.
% The AdamW~\cite{Loshchilov2017DecoupledWD} optimizer with a weight decay was used with a learning rate of 0.001. 
Experiments were conducted with GPT2-Medium, FLAN-T5 Large and Llama 3.2 3B Instruct models. 
{\gpt} and {\flan} were fine-tuned fully, while {\llamai} used Low-Rank Adaptation~(LoRA)~\cite{Hu2021LoRALA} and 8-bit quantization~\cite{jacob2018quantization} for memory efficiency.

\vspace{-3pt}
\section{Experimental Setup}
\vspace{-4pt}

% \noindent
% \textbf{Datasets.}
\subsection{Datasets}
We use two ToD datasets: Schema-Guided Dialog (SGD) dataset, and Knowledge-Enhanced Task-Oriented Dialog (KETOD) dataset. 
Table~\ref{tab:data_statistics} shows detailed statistics about the datasets.
These datasets are publicly available, large, and represent a wide range of domains that span different tasks.
We have selected these datasets as they describe the domain using schema and have the necessary information to simulate communication with external resources through {\apicall}s.

\begin{table}
    \centering
    \small
    \begin{adjustbox}{max width=0.49\textwidth}
        \begin{tabular}{lcc}
            \hline
            % Datasets $\rightarrow$ & \textbf{SGD} & \textbf{KETOD} & \textbf{BiToD}  \\ \hline
            % \# Dialogs & 16,142  & 5,324 & 3,689 \\ \hline
            % Average Turns / Dialog & 20.44   & 9.78 & 9.39 \\ \hline
            % \# Domains & 20 & 14 & 5 \\ \hline
            % \# Seen Domains & 16 & 12 & 5 \\ \hline
            % \# Unseen Domains & 4 & 2 & 0 \\ \hline
            % \# Unique API methods  & 46  & 38 &  8 \\ \hline
            % \# Unique API parameters & 214 &  195 & 20 \\ \hline
            Datasets & \textbf{SGD} & \textbf{KETOD} \\ %\hline
            \hline
            \# Dialogs & 16142  & 5324 \\ %\hline
            Average Turns / Dialog & 20.44   & 9.78 \\ % \hline
            % \# Domains & 20 & 14 \\ \hline
            % \# Seen Domains & 16 & 12 \\ \hline
            % \# Unseen Domains & 4 & 2 \\ \hline
            % \# {\apicall}s & 13239 & - \\
            \# Unique API methods:all & 46  & 46 \\ 
            \# Unique API methods:unseen  & 8  & 8 \\ 
            \# Unique API parameters: all& 137 & 134 \\
            \# Unique API parameters: unseen & 88 & 88 \\
            \hline
            % \# Unique API parameters & 214 &  195 \\ \hline
            
        \end{tabular}
    \end{adjustbox}
    \vspace{-8pt}
    \caption{Dataset statistics.}
    \label{tab:data_statistics}
    \vspace{-20pt}
\end{table}

% \noindent
% \textbf{Evaluation.}
\vspace{-3pt}
\subsection{Evaluation}

We evaluate the system across four domain categories: \textit{All Domains} (dialogs from all domains), \textit{Seen Domains} (dialogs from training domains), \textit{Unseen Domains} (dialogs from domains not included in the training data), and \textit{Mixed Domains} (dialogs with both seen and unseen domains).
% To better understand the generalization ability of {\oursys}, we evaluate the system on dialogs grouped by domain categories. This evaluation is structured into for categories: All, Seen, Unseen, and Mixed.
% \textit{All Domains} includes dialogs from all domains, providing a comprehensive evaluation of the system's overall performance.
% \textit{Seen Domains} contains dialogs from domains that were present in the training data. This setting demonstrates the supervised learning performance.
% \textit{Unseen Domains} comprises of dialogs from domains that were not included in the training data. This setting demonstrates the out-of-domain generalization performance.
% \textit{Mixed Domains} consists of dialogs that have multiple domains, where some domains are in seen and some in unseen. This setting serves as an intermediary between supervised and out-of-domain setting. It tests the system's ability to manage dialogs where known and unknown domains are intermixed, reflecting more complex real-world scenarios.
We analyze the performance of overall responses as well as its sub-tasks---Request and Inform. For task completion, we introduce custom metrics to assess the performance of individual components.
% We report scores across four domain categories: \textit{All}, \textit{Seen}, \textit{Unseen}, and \textit{Mixed}. \textit{All Domains} includes dialogs from all domains, offering a comprehensive performance evaluation. \textit{Seen Domains} consists of dialogs from training domains, reflecting supervised learning performance, while \textit{Unseen Domains} contains dialogs from domains not included in the training data, testing out-of-domain generalization. Mixed Domains features dialogs with both seen and unseen domains, testing the system's ability to manage mixed real-world scenarios.
% We evaluate {\oursys} on its overall responses as well as its performance on the sub-tasks—Request and Inform—using custom metrics to assess task completion and component performance. 

% \fixed{: I have moved it to the correct position. }\fix{I do not see metrics here.}

\vspace{-3pt}
\begin{table*}[!t]
    \centering
    \begin{adjustbox}{max width=\textwidth}
    \small
        \begin{tabular}{|c|c|c|c c c c| c c c c|}
            \hline
    \multirow{2}{*}{\textbf{Dataset}} & \multirow{2}{*}{\textbf{Model}} & \textbf{Annotations} &  \multicolumn{4}{c|}{\textbf{Overall Response (BertScore-F1)}} &  \multicolumn{4}{c|}{\textbf{Complete API Accuracy}} \\ %\cline{4-11}
    & & \textbf{Required}& \textbf{all}  & \textbf{seen} & \textbf{mixed} & \textbf{unseen} & \textbf{all}  & \textbf{seen} & \textbf{mixed} & \textbf{unseen} \\ \hline
    \multirow{6}{*}{SGD} 
    & \simpletod & Yes & 0.6100 & 0.600 & 0.6300 & 0.5800 
    & 20.30 & 44.94 & 25.57 & 09.39 \\ 
    & \soloist & Yes & 0.6214 & 0.6538 & 0.6265 & 0.6097 
    & 19.25 & 47.82 & 24.32 & 07.66 \\ 
    & \zstod & Yes & 0.5704 & 0.6439 & 0.5648 & 0.5600 
    & 20.38 & 56.15 & 20.28 & 12.62 \\ \cline{2-11}
    & \gpt & No & 0.7002 & 0.7291 & 0.7149 & 0.6800
    & 35.53 & 74.77 & 42.87 & 19.24 \\ 
    & \llamai & No & \underline{0.7629} & \textbf{0.7850} & \underline{0.7708} & \underline{0.7507}
    & \underline{52.84} & \textbf{90.19} & \underline{57.44}  & \underline{39.84}\\
    & \flan & No & \textbf{0.7633} & \underline{0.7792} & \textbf{0.7723} & \textbf{0.7513}
    & \textbf{65.87} & \underline{89.88} & \textbf{72.99} & \textbf{53.16}  \\ 
    % \thickhline
    \hline
    \multirow{6}{*}{KETOD}
    & \simpletod & Yes & 0.5248 & 0.5540 & 0.5382 & 0.4735
    & 36.24 & 61.07 & 31.19 & 08.74 \\
    & \soloist & Yes & 0.5035 & 0.5201 & 0.4895 & 0.4983 
    & 24.12 & 43.79 & 17.29 & 05.98  \\
    & \zstod & Yes & 0.4759 & 0.4822 & 0.4643 & 0.4809
    & 26.70 & 43.29 & 21.75 & 10.34 \\ \cline{2-11}
    & \gpt & No & 0.6766 & 0.7001 & 0.6821 & 0.6410
    & 36.75 & 59.56 & 32.62  & 10.80\\
    & \llamai & No & \underline{0.7369} & \underline{0.7624} & \underline{0.7363} & \underline{0.7057}
    & \underline{63.32} & \underline{91.61} & \underline{55.08}  & \underline{35.17} \\
    & \flan & No & \textbf{0.7431} & \textbf{0.7665} & \textbf{0.7457} & \textbf{0.7112}
    % & \underline{59.11} & \underline{79.19} & \textbf{55.97} & \textbf{35.63}\\ 
    & \textbf{77.26} & \textbf{95.97} & \textbf{75.58}  & \textbf{53.79} \\ 
    \hline
        \end{tabular}
    \end{adjustbox}
    \vspace{-6pt}
    % \caption{RQ 1: Turn annotations.}
    % \caption{Experimental results for RQ1 evaluated on Overall Response Generation and Complete API Accuracy metrics.}
    % \caption{Evaluating the impact of Data Annotations on Model Performance for Response Generation and API Accuracy (RQ1).}
    \caption{Performance comparison between annotation-dependent baselines and annotation-free {\oursys} models on Overall Response and API Accuracy (RQ1).}
    \label{tab:turn_annotations}
    \vspace{-6pt}
\end{table*}

\vspace{4pt}
\noindent
\textbf{Response Generation.}
To evaluate the quality of the response generation of models, we report BERTScore.
We used \texttt{microsoft/mpnet-base} as the model type for calculating the BERTScore. We report BLEU-4~\cite{Papineni2002BleuAM} scores in Appendix~\ref{sec:appendix_bleu}

% \fixed{: Reorganized some text. }\fix{I think, the next items are old, please revise.}
\noindent
\textbf{{\apicall}s.} 
The format for an {\apicall} is:
\(
\texttt{APICall}( \texttt{method=method\_name, parameters} = \{ (s_i,v_i)_{i=1}^n \} ).
\)
The parameters attribute is a list of slot name and slot value pairs, where $s_i$ represents the slot name and $v_i$ represents the value of that slot. 

We use regular expressions to extract different parts of the {\apicall}, and apply custom metrics to access different parts of an {\apicall}.

\textit{Invoke Accuracy} measures whether the system can understand when to make an \apicall.
%metric ensures that the correct type of API query was performed by verifying the query type. This metric also measures whether the system can understand when to make an API call.    
 \textit{Method Accuracy} checks whether the appropriate method name was used in the \apicall.      
 \textit{Param Name Accuracy} assesses whether all the parameter names used to construct the {\apicall} are accurately.
 \textit{Param Value Accuracy} evaluates whether each parameter value corresponding to a parameter name is correct. It is important to note that this metric will only be considered if the corresponding parameter name is correct. 
 %If the parameter name is incorrect, the parameter value accuracy is scored as 0, even if the value itself was correct. Instead of performing an exact string match, fuzzy string matching is performed.
 \textit{Complete {\apicall} Accuracy} metric checks whether the complete {\apicall} (i.e., all components) was generated correctly.

\begin{table*}[!t]
    \centering
    \begin{adjustbox}{max width=\textwidth}
    \small
        \begin{tabular}{|c|c|c|cccc| cccc|}
            \hline
    \multirow{2}{*}{\textbf{Dataset}} & \multirow{2}{*}{\textbf{Model}} & \textbf{Annotations} &  \multicolumn{4}{c|}{\textbf{Inform (BertScore-F1)}} &  \multicolumn{4}{c|}{\textbf{Request (BertScore-F1)}} \\ %\cline{4-11}
    & & \textbf{Required}& \textbf{all}  & \textbf{seen} & \textbf{mixed} & \textbf{unseen} & \textbf{all}  & \textbf{seen} & \textbf{mixed} & \textbf{unseen} \\ \hline
    \multirow{6}{*}{SGD} 
    & \simpletod & Yes & 0.6100 & 0.6400 & 0.6300 & 0.5800 
    & 0.4300 & 0.4000 & 0.4500 & 0.4200 \\ 
    & \soloist & Yes & 0.6596 & 0.6982 & 0.6730 & 0.6356 
    & 0.4852 & 0.5069 & 0.4797 & 0.4858 \\ 
    & \zstod & Yes & 0.4714 & 0.5324 & 0.4590 & 0.4681 
    & 0.5012 & 0.5561 & 0.4944 & 0.4970 \\ \cline{2-11}
    & \gpt & No & 0.7451 & 0.7597 & 0.7516 & 0.7344 
    & 0.5287 & 0.5200 & 0.5302 & 0.5291 \\ 
    & \llamai & No & \textbf{0.7853} & \textbf{0.7962} & \textbf{0.7904} & \underline{0.7771} 
    & \textbf{0.6073} & \textbf{0.6244} & \textbf{0.6038} & \textbf{0.6071} \\ 
    & \flan & No & \underline{0.7838} & \underline{0.7906} & \underline{0.7876} & \textbf{0.7781} 
    & \underline{0.6034} & \underline{0.6169} & \underline{0.5965} & \underline{0.6066}  \\ \thickhline
    \multirow{6}{*}{KETOD}
    & \simpletod & Yes & 0.4275 & 0.4445 & 0.4637 & 0.3659
    & 0.4267 & 0.4341 & 0.4376 & 0.4076 \\
    & \soloist & Yes & 0.5076 & 0.4961 & 0.5317 & 0.4971
    & 0.4638 & 0.4588 & 0.4604 & 0.4729 \\
    & \zstod & Yes & 0.3489 & 0.3334 & 0.3456 & 0.3734
    & 0.5167 & 0.5473 & 0.4989 & 0.4991 \\ \cline{2-11}
    & \gpt & No & 0.6831 & 0.7048 & 0.6878 & 0.6489
    & 0.4814 & 0.4919 & 0.4733 & 0.4772 \\
    & \llamai & No & \textbf{0.7488} & \textbf{0.7691} & \underline{0.7503} & \textbf{0.7198}
    & \underline{0.5976} & \textbf{0.6198} & \underline{0.5794} & \textbf{0.5901} \\
    & \flan & No & \underline{0.7447} & \underline{0.7656} & \textbf{0.7607} & \underline{0.6994}
    & \textbf{0.5981} & \underline{0.6189} & \textbf{0.5893} & \underline{0.5829}  \\ \hline    
        \end{tabular}
    \end{adjustbox}
    \vspace{-6pt}
    % \caption{RQ 1: Additional Response Metrics.}
    \caption{Results for Response Generation sub-tasks: Inform and Request (RQ1).}
    \label{tab:additional_response}
    \vspace{-15pt}
\end{table*}

\vspace{-5pt}
\subsection{Baselines}
\vspace{-3pt}
% To measure the effectiveness of our approach, we compare {\oursys} against popular SOTA approaches. 
% It is important to note that, apart from {\autotod}, all other approaches use annotated data. Existing approaches do not report {\apicall} metrics, so we implemented them to the best of our ability to report the metrics. {\soloist}, {\simpletod} and {\zstod} were all implemented using a GPT-2 Medium model. For these approaches, during inference we extract the system response from the generation and disregard additional information like the dialog state and system actions.
\noindent
\textit{\soloist}~\cite{Peng2021SoloistBT} introduced an E2E ToD system that employs a transformer-based autoregressive model that generates dialog responses grounded in user goals and real-world knowledge for task completion.

\noindent
\textit{SimpleTOD}~\cite{Chen2022KETODKT} introduced a ToD model as an end-to-end sequence generation problem that utilizes the dialog history, dialog states and system actions to generate system responses.

\noindent
\textit{ZS-TOD}~\cite{Mosharrof2023ZeroShotGE} introduced a zero-shot generalizable E2E ToD model that incorporates domain schema and dialog annotations to generate dialog responses.

\noindent
\textit{AutoTOD}~\cite{Xu2024RethinkingTD} introduced a zero shot autonomous ToD agent, that works without manual annotations and also has the ability to communicate with external resources.

{\soloist}, {\simpletod} and {\zstod} were implemented using GPT-2 Medium. During inference, we extract the system response and disregard the additional information like dialog state and system actions.
\vspace{-9pt}
\section{Results}
\vspace{-3pt}

% \fix{I do not see proper captions of tables/tables.}

\vspace{-5pt}
\noindent
% \textbf{Performance without turn level annotations.}
% \begin{tcolorbox}[colframe=gray!20, colback=gray!10, coltitle=black, boxrule=0.4mm, width=\columnwidth, left=1mm, right=1mm, top=1mm, bottom=1mm]
% RQ1: Can TOD systems operate without annotated data? 
% \end{tcolorbox}
\noindent
%Table~\ref{tab:turn_annotations} presents the findings related to RQ1: Can TOD systems operate without annotated data? 
Table~\ref{tab:turn_annotations} presents the findings for \textit{RQ1: Can pre-trained LLMs be adapted into effective ToD systems without turn-level annotated data ?}
Our results show that {\oursys} models, which do not rely on turn-level annotations, outperform models trained with annotated data in response generation.
A key reason for this improvement is that {\oursys} models focus solely on generating system responses, whereas annotation-based models must produce structured outputs that include dialog state, system actions, and responses—requiring the model to optimize for multiple complex tasks simultaneously. 
Furthermore, the substantial performance gap between the baseline approaches built with {\gpt} and the {\gpt} variant of {\oursys} suggests that learning to generate responses directly is a more effective approach for ToD systems.

For task completion, all models trained without turn-level annotations consistently outperform the annotated models. This finding highlights the sufficiency of dialogue history as a standalone source of context for completing complex tasks. 
Table~\ref{tab:turn_annotations} reveals more insights about the different {\oursys} models.  {\flan} and {\llamai} being the larger models, significantly outperform the smaller {\gpt} model for task completion. 
However, even though {\llamai} is a larger model than {\flan}, it does not have better task completion performance. 

This discrepancy may stem from differences in the training methodologies. Specifically, {\llamai} was trained using 8-bit quantization and LoRA adapters, whereas {\flan} underwent full fine-tuning. The use of LoRA significantly reduces the number of trainable parameters and the 8-bit quantization introduces precision loss due to the reduced bit width. These factors likely contributed to {\llamai}'s lower performance despite its larger model size.

% \fixed{: Added text about lora and 8bit. }\fix{above passage:
% Size of llama? is it lora? also 8-bit?}

\vspace{-3pt}
\begin{table*}[!t]
    \centering
    \begin{adjustbox}{max width=\textwidth}
    \small
        \begin{tabular}{|c|c|c|cccc| cccc|}
            \hline
    \multirow{2}{*}{\textbf{Dataset}} & \multirow{2}{*}{\textbf{Model}} & \textbf{Augm-}  & \multicolumn{4}{c|}{\textbf{Overall Response(BertScore-F1)}} &  \multicolumn{4}{c|}{\textbf{Complete API Accuracy}} \\ %\cline{4-11}
    & & \textbf{ented} & \textbf{all}  & \textbf{seen} & \textbf{mixed} & \textbf{unseen} & \textbf{all}  & \textbf{seen} & \textbf{mixed} & \textbf{unseen} \\ \hline
    \multirow{6}{*}{SGD} 
    & \gpt & \xmark & 0.7002 & 0.7291 & 0.7149 & 0.6800 
    & 35.53 & 74.77 & 42.87  & 19.24 \\ 
    & \gpt & \cmark & 0.7266 & 0.7431 & 0.7437 & 0.7068 
    & 47.66 & 82.01 & 53.76 & 33.75 \\ %\cline{2-11}
    & \llamai & \xmark & \underline{0.7629} & \underline{0.7850} & \underline{0.7708} & 0.7057
    & 52.84 & 90.19 & 57.44  & 39.84\\ 
    & \llamai & \cmark & 0.7623 & \textbf{0.7852} & 0.7693 & \underline{0.7506}
    & 62.38 & \textbf{94.31} & 63.99  & \underline{53.68} \\ %\cline{2-11}
    & \flan & \xmark & \textbf{0.7633} & 0.7792 & \textbf{0.7723} & \textbf{0.7513}
    & \underline{65.87} & 89.88 & \underline{72.99}  & 53.16 \\ 
    & \flan & \cmark & 0.7320 & 0.7494 & 0.7411 & 0.7196
    & \textbf{73.49} & \underline{90.65} & \textbf{81.76}  & \textbf{61.07} \\ 
    % \thickhline
    \hline
    \multirow{6}{*}{KETOD}
    & \gpt & \xmark & 0.6766 & 0.7001 & 0.6821 & 0.6410
    & 36.75 & 59.56 & 32.62  & 10.80\\
    & \gpt & \cmark & 0.6677 & 0.6867 & 0.6738 & 0.6372
    & 48.43 &72.48 & 40.64  & 25.52\\ %\cline{2-11}
    & \llamai & \xmark & 0.7369 & 0.7624 & 0.7363 & 0.7057
    & 63.32 & 91.61 & 55.08  & 35.17 \\ 
    & \llamai & \cmark & 0.7405 & \underline{0.7679} & 0.7389 & 0.7082
    & 73.24 & \textbf{97.48} & 67.02 & 48.05 \\ %\cline{2-11}
    & \flan & \xmark & \underline{0.7431} & 0.7665 & \underline{0.7457} & \underline{0.7112}
    % & 59.11 & 79.19 & 55.97  & 35.63 \\ 
    & \underline{77.26} & 95.97 & \underline{75.58}  & \underline{53.79} \\ 
    & \flan & \cmark & \textbf{0.7549} & \textbf{0.7786} & \textbf{0.7541} & \textbf{0.7261}
    % & \textbf{76.26} & 90.10 & \textbf{74.51}  & \textbf{59.54} \\ \hline
    & \textbf{82.66} & \underline{96.48} & \textbf{82.35}  & \textbf{64.14} \\ \hline
        \end{tabular}
    \end{adjustbox}
    \vspace{-6pt}
    % \caption{RQ 2: Schema Augmentations}
    \caption{Impact of Schema Augmentation Mechanism on Response Generation and API Accuracy (RQ2).}
    \label{tab:augmentation}
    \vspace{-8pt}
\end{table*}

\noindent
\textbf{Detailed Response Generation Performance.}
To get a better understanding of the response generation task, we break it down into two sub-tasks---Inform and Request---and present the results in Table~\ref{tab:additional_response}. The inform sub-task focuses on providing responses to user requests, while the request sub-task involves prompting users for additional information. 
Similar to Table~\ref{tab:turn_annotations}, the results here show a consistent trend, with {\oursys} models outperforming those trained on annotated data.
Additionally, we observe that the Request sub-task is significantly more challenging than Inform. This is expected, as there are multiple plausible pieces of information a system could request, but if they do not align with the gold standard, the model receives a lower score. In contrast, the Inform sub-task is more straightforward since the user explicitly requests specific information, making it easier for the system to generate the correct response.

\noindent
\textbf{Schema Augmentation Performance.}
Table~\ref{tab:augmentation} presents the results for \textit{RQ2: How can we improve the {\ood} generalization of ToD systems for task completion??} 
% It compares the performance of {\oursys} models with the schema augmentation mechanism. 
Across all the models, we can see that the response generation performance is similar, but there are improvements in task completion performance, specially a big increment in the unseen domain. For seen domains, there is a small improvement, which is expected as the augmentation mainly teaches the models how to use the schema to generalize to {\ood} data, however for unseen domains, this learning is very useful and the models have shown considerable improvements.
Between {\llamai} and {\flan}, we can see that for seen domains {\llamai} has a slightly better performance however for unseen domains {\llamai} has much lower performance. One reason for this could be the size of the two models, {\llamai} being the larger model may have a higher capacity to memorize the training data, which could explain its stronger performance on seen domains. However, this can also make it more prone to over-fitting and may not generalize well to new, unseen domains.

\vspace{-3pt}
\begin{table*}[!t]
    \centering
    \begin{adjustbox}{max width=\textwidth}

        \begin{tabular}{|c|c|c|c c c| c c c| c c c| c c c|}
            \hline

        \multirow{2}{*}{\textbf{Dataset}} & \multirow{2}{*}{\textbf{Model}} & \textbf{Augm-}   & \multicolumn{3}{c|}{\textbf{API Invoke Accuracy}} & \multicolumn{3}{c|}{\textbf{API Method Accuracy}} & \multicolumn{3}{c|}{\textbf{Param Names Accuracy}} & \multicolumn{3}{c|}{\textbf{Param Values Accuracy}} 
    \\ % \cline{3-15}
    & & \textbf{ented} & \textbf{all}  & \textbf{seen} & \textbf{unseen}
    & \textbf{all}  & \textbf{seen} & \textbf{unseen}
    & \textbf{all}  & \textbf{seen} & \textbf{unseen}
    & \textbf{all}  & \textbf{seen} & \textbf{unseen}
    \\ \hline
    \multirow{9}{*}{SGD} 
    & \soloist & \xmark & 79.92 & 80.37 & 79.44 
    & 64.51 & 72.82 & 60.74 
    & 47.06 & 72.05 & 35.54 
    & 45.16 & 71.30 & 33.13 
    \\
    & \simpletod & \xmark & 66.28 & 54.52 & 68.44
    & 32.94 & 26.32 & 33.53 
    & 28.52 & 29.13 & 26.25 
    & 26.98 & 28.54 & 24.45
    \\
    & \zstod & \xmark & 86.65 & 90.97 & 86.91
    & 64.18 & 80.69 & 61.82
    & 44.44 & 73.15 & 40.17
    & 42.74 & 72.22 & 38.13
    \\ \cline{2-15}
    & \gpt & \xmark & 90.51 & 97.66 & 86.21
    & 78.62 & 96.26 & 71.42
    & 60.65 & 92.07 & 49.52
    & 58.64 & 91.24 & 46.98
    \\
    & \gpt & \cmark & 93.89 & 96.81 & 91.26
    & 83.07 & 89.17 & 77.92 
    & 76.94 & 91.71 & 72.15
    & 74.18 & 90.86 & 68.59
    \\ %\cline{2-15}
    & \llamai & \xmark & 98.08 & \textbf{99.69} & \underline{97.52} 
    & 92.18 & \textbf{99.69} & 91.00
    & 84.53 & \underline{98.36} & 80.44
    & 81.55 & \underline{97.92} & 76.55
    \\
    & \llamai & \cmark & \textbf{98.78} & 99.38 & \textbf{98.80}
    & 95.26 & 99.38 & \underline{94.97}
    & \underline{89.23} & \textbf{99.10} & \underline{88.11}
    & \underline{86.25} & \textbf{98.73} & \underline{84.02}
    \\ %\cline{2-15}
    & \flan & \xmark & \underline{98.65} & \underline{99.61} & 98.24
    & \textbf{96.84} & \underline{99.61} & \textbf{95.23}
    & 79.71 & 96.97 & 74.58 
    & 76.03 & 94.04 & 70.43
    \\
    & \flan & \cmark & 96.28 & 99.53 & 94.42
    & \underline{95.97} & 99.53 & 93.89 
    & \textbf{94.37} & 97.23 & \textbf{92.17}
    & \textbf{92.08} & 96.99 & \textbf{88.86}
    \\ 
    \thickhline
    % \hline
    \multirow{9}{*}{KETOD}
    & \soloist & \xmark & 52.20 & 56.04 & 44.14
    & 42.96 & 47.65 & 34.25
    & 37.10 & 47.43 & 24.94
    & 36.31 & 46.98 & 23.95
    \\
    & \simpletod & \xmark & 44.85 & 41.95 & 41.84 
    & 34.30 & 33.39 & 29.43
    & 25.87 & 30.74 & 16.87
    & 25.31 & 30.38 & 16.30
    \\
    & \zstod & \xmark & 43.03 & 43.62 & 40.46
    & 32.91 & 36.58 & 31.72
    & 28.06 & 31.92 & 26.89
    & 26.53 & 30.94 & 24.43
    \\ \cline{2-15}
    & \gpt & \xmark & 78.83 & 80.87 & 73.56
    & 71.48 & 78.36 & 62.53
    & 57.02 & 72.98 & 39.60
    & 55.46 & 71.83 & 37.73
    \\ 
    & \gpt & \cmark & 92.96 & 91.78 & 94.48
    & 86.62 & 88.09 & 85.29
    & 75.66 & 84.53 & 69.97
    & 72.85 & 83.55 & 65.02
    \\ %\cline{2-15}
    & \llamai & \xmark & \underline{96.55} & \underline{97.48} & \underline{95.17} 
    & \underline{92.90} & \underline{96.98} & 88.97
    & 86.04 & \underline{97.13} & 76.43
    & 84.26 & \underline{96.65} & 73.37
    \\
    & \llamai & \cmark & \textbf{98.49} & \textbf{99.16} & \textbf{97.70}
    & \textbf{96.80} & \textbf{99.16} & \textbf{94.94}
    & \textbf{91.86}  & \textbf{98.67} & \underline{86.05}
    & \textbf{90.10} & \textbf{98.62} & \underline{82.83} 
    \\ %\cline{2-15}
    & \flan & \xmark & 90.45 & 89.93 & 90.34
    & 89.07 & 89.93 & 86.67  
    & 78.62 & 88.92 & 67.93
    & 76.52 & 87.94 & 65.28
    \\ 
    & \flan & \cmark & 92.34 & 93.46 & 92.18
    & 92.15 & 93.12 & \underline{92.18}
    & \underline{90.77} & 93.53 & \textbf{88.84}
    & \underline{88.86} & 93.23 & \textbf{85.26}
    \\ \hline
        \end{tabular}
    \end{adjustbox}
    \vspace{-6pt}
    % \caption{Additional API Metrics for Baselines and {\oursys} models with and without augmentation (RQ1, RQ2).}
    \caption{Additional API Metrics for baseline approaches and {\oursys} models (RQ1, RQ2).}
    \label{tab:api_metrics}
    \vspace{-8pt}
\end{table*}

\noindent
\textbf{Detailed Task Completion Performance.}
To complete a task, a model has to make a correct {\apicall}. An {\apicall} has many aspects in it, and we present detailed results in Table~\ref{tab:api_metrics}. We can see that {\oursys} models considerably outperform baseline approaches across all metrics. Upon inspecting the {\apicall} Invoke Accuracy, we see that baseline approaches have much lower scores, indicating that they struggle in identifying when to make {\apicall}s. The {\apicall} Method Accuracy evaluates whether a model generates the correct method name in the \apicall. A common pattern that we see across all models is that there is a drop in parameter names accuracy when compared to the previous metrics. Generating the correct list of parameters for the {\apicall} is inherently a harder problem than deciding when to make an {\apicall} and what method to use, so the performance degradation is understandable. 

A key observation from Table~\ref{tab:api_metrics} is the significant impact of the schema augmentation on the {\apicall} parameter names metric. Our results indicate that schema augmentation yields the largest improvement for this metric. {\apicall} parameters are directly derived from the schema, and schema augmentation enables the models to better recognize and utilize these patterns, thus improving the model's ability to generate the correct list of parameters, leading to a notable increase in parameter names accuracy. Furthermore, the {\apicall} parameter values accuracy also improved as a result, since a model is only rewarded for generating the correct value if it is assigned to the appropriate parameter name. 

For instance, consider the task of finding a bus using the \texttt{FindBus} method. We compare two schema variations, \texttt{Buses\_1} and \texttt{Buses\_11}, which define different slot names for the same concepts. In \texttt{Buses\_1}, the slot names are \texttt{from\_station} and \texttt{to\_station}, and for \texttt{Buses\_11}, the slot names are \texttt{origin} and \texttt{destination}.

A model trained without schema augmentation tends to overfit to specific slot names seen during training. If the model was trained on \texttt{Buses\_1}, it might always generate \texttt{from\_station} and \texttt{to\_station}, even when interacting with \texttt{Buses\_11}, leading to incorrect {\apicall}s.
For example, given the user utterance: \textit{``I want to find a bus from LA to SFO''}, the model without augmentation might generate: 
\(
\begin{aligned}
    &\quad\texttt{\apicall(method=FindBus, parameters=} \\
    &\quad \quad \texttt{from\_station=LA, to\_station=SFO \}).}
\end{aligned}
\)
% $$
% \texttt{\apicall(method=FindBus, parameters=\{from\_station=LA, to\_station=SFO \}).} 
% $$
In the \texttt{Buses\_11} schema, the slot names \texttt{from\_station} and \texttt{to\_station} do not exist, thus making the {\apicall} invalid.

On the other hand, a model trained with schema augmentation learns to generalize across schema variations by recognizing slot name patterns from multiple schemas, and might generate:
\(
\begin{aligned}
    &\quad\texttt{\apicall(method=FindBus, parameters=} \\
    &\quad\quad\texttt{origin=LA, destination=SFO\}).} 
\end{aligned}
\)
% $$
% \texttt{\apicall(method=FindBus, parameters=\{departure\_station=LA, arrival\_station=SFO\}).}
% $$

The model can dynamically align its output with the schema it is conditioned on. By learning to use the slot names from the provided schema rather than relying on the memorized slot names, a model trained with schema augmentation demonstrates improved robustness and generalization.

% \fixed{: Added example with schema variations. }\fix{The above needs a better explanation about generalization through schema. Also, we should not bring Washington, D.C. and DC (a simple look-up can fix it.)}

\noindent
\textbf{Fine-tuning Performance.}
% Table~\ref{tab:autotod_results} presents the results on unseen domains for {\autotod} and {\oursys} models with schema augmentation. 
% Table~\ref{tab:autotod_results} presents the results for \textit{RQ3: How does the {\ood} generalization of fine-tuned ToD systems compare to that of large-scale, proprietary LLMs?} by comparing the performance of {\oursys} models against {\autotod}, which was built using GPT-4o.
Table~\ref{tab:autotod_results} presents the results on unseen domains for {\oursys} models, and {\autotod}, which was built using GPT-4o. Using the results in Table~\ref{tab:autotod_results}, we can answer \textit{RQ3: How does the {\ood} generalization of fine-tuned ToD systems compare to that of large-scale, proprietary LLMs?}
For the Complete API Accuracy metric, except for the {\gpt} model, all other {\oursys} models outperform {\autotod}. For all the other metrics, {\autotod} has much lower scores than the {\oursys} models. 
A key metric to note here is the API Invoke Accuracy, which measures whether a model is making an API call on the right turn, and {\autotod} has a very low score on this metric when compared to {\oursys} models.
Due to this issue, {\autotod} also has a much lower score for the Overall Response metric, as it makes \apicall s on turns where a general interaction is expected. Based on these results, we can state that fine-tuning is an important step to identify the timing of making an {\apicall} in ToD systems.

\begin{table*}[!t]
    \centering
    \small
       \begin{adjustbox}{max width=\textwidth}
        \begin{tabular}{|c|l|c c c c| c c c c|}
            \hline
            \multirow{2}{*}{\textbf{Dataset}} & \multirow{2}{*}{\textbf{Domains}} & \multicolumn{4}{c|}{\textbf{API Invoke Accuracy}} & \multicolumn{4}{c|}{\textbf{Complete API Accuracy}} \\
            % \cline{3-10} 
            % \cline{11-18}
            & & \textbf{Auto-ToD} & \textbf{\gpt} & \textbf{\llamai} & \textbf{\flan} & \textbf{Auto-ToD} & \textbf{\gpt} & \textbf{\llamai} & \textbf{\flan} \\ 
            \hline
\multirow{9}{*}{SGD} & Alarm\_1       & 77.78 & 71.11 & \underline{98.89} & \textbf{100.00} & \underline{76.67} & 15.56 & \textbf{78.89} & 61.11 \\
& Buses\_3       & 62.84 & 94.59 & \textbf{100.00} & \underline{99.32} & 37.16 & 29.05 & \underline{46.62} & \textbf{57.43} \\
& Events\_3      & 64.29 & 92.06 & \underline{92.86} & \textbf{96.83} & 27.78 & \underline{55.56} & 50.79 & \textbf{60.32} \\
& Homes\_2       & 43.06 & 97.92 & \textbf{100.00} & \underline{99.31} & 39.58 & 69.44 & \textbf{76.39} & \underline{74.31} \\
& Hotels\_4      & 43.88 & 94.24 & \textbf{100.00} & \underline{98.56} & 40.29 & 49.64 & \textbf{100.00} & \underline{77.70} \\
& Movies\_3      & 67.80 & 54.24 & \textbf{98.31} & \underline{96.61} & 47.46 & 23.73 & \textbf{77.97} & \underline{67.80} \\
& Music\_3       & 41.51 & 94.34 & \underline{98.11} & \textbf{100.00} & 30.19 & 58.49 & \underline{73.58} & \textbf{84.91} \\
& RentalCars\_3  & 48.67 & \underline{99.12} & \underline{99.12} & \textbf{100.00} & 41.59 & 35.40 & \underline{54.87} & \textbf{63.72} \\
& Restaurants\_2 & 60.29 & 99.26 & \textbf{100.00} & \textbf{100.00} & 28.68 & \underline{84.56} & 77.94 & \textbf{85.29} \\
% \thickhline
\hline
 \multirow{9}{*}{KETOD} & Alarm\_1       & 66.67 & \textbf{100.00} & \textbf{100.00} & \textbf{100.00} & \underline{66.67} & 00.00 & \textbf{100.00} & \underline{66.67} \\
 & Buses\_3       & 50.00 & 86.36 & \underline{95.45} & \textbf{100.00} & 09.09 & 18.18 & \textbf{45.45} & \underline{36.36} \\
 & Events\_3      & 81.82 & \textbf{100.00} & \textbf{100.00} & 90.91 & 45.45 & \textbf{72.73} & \textbf{72.73} & 54.55 \\
 & Homes\_2       & 70.59 & \textbf{100.00} & \underline{94.12} & 93.75 & \textbf{70.59} & 41.18 & 64.71 & \textbf{70.59} \\
 & Hotels\_4      & 43.75 & \textbf{100.00} & \textbf{100.00} & \textbf{100.00} & 37.50 & 37.50 & \underline{81.25} & \textbf{100.00} \\
 & Movies\_3      & 71.43 & \underline{85.71} & \underline{85.71} & \textbf{95.00} & \textbf{57.14} & 28.57 & \textbf{57.14} & \textbf{57.14} \\
 & Music\_3       & 22.22 & \underline{88.89} & \textbf{100.00} & 78.26 & 22.22 & 44.44 & \textbf{88.89} & \underline{66.67} \\
 & RentalCars\_3  & 62.50 & \textbf{100.00} & \textbf{100.00} & \textbf{100.00} & \textbf{56.25} & 31.25 & \underline{50.00} & \underline{50.00} \\
 & Restaurants\_2 & 50.00 & \textbf{100.00} & \underline{87.50} & \underline{87.50} & 25.00 & 00.00 & \textbf{87.50} & \underline{62.50} \\
 \hline
        \end{tabular}
    \end{adjustbox}
    \vspace{-6pt}
    \caption{Domain-wise evaluation of API Invoke and Complete API on unseen domains for {\autotod} and {\oursys} models (RQ3).}
    \label{tab:unseen_domain_results}
    \vspace{-10pt}
\end{table*}

\noindent
\textbf{Domain Specific Results.}
To get a deeper understanding of the performance of {\autotod} and {\oursys} models, we present some domain specific results for the API Invoke Accuracy and Complete API Accuracy metrics in Table~\ref{tab:unseen_domain_results}. For the Api Invoke Accuracy, we see the same pattern as before, with {\autotod} having much lower scores than {\oursys} models. 
From these results, we can make another interesting observation, {\autotod} has higher Complete Api Accuracy for simple domains like \texttt{Alarm} and \texttt{Movies}, however it has poor performance for complex domains like \texttt{Restaurants}, \texttt{Buses}, and \texttt{Music}. 
Since {\oursys} models have been fine-tuned, the models have a better understanding of the structure of complex domains. The models do not have a big drop in performance across domains, showing the robustness achieved through fine-tuning. {\flan} and {\llamai} being the larger models, show more stability in performance over the smaller {\gpt} model.

\vspace{-5pt}
\begin{figure}
   \centering
   \includegraphics[width=0.95\linewidth]{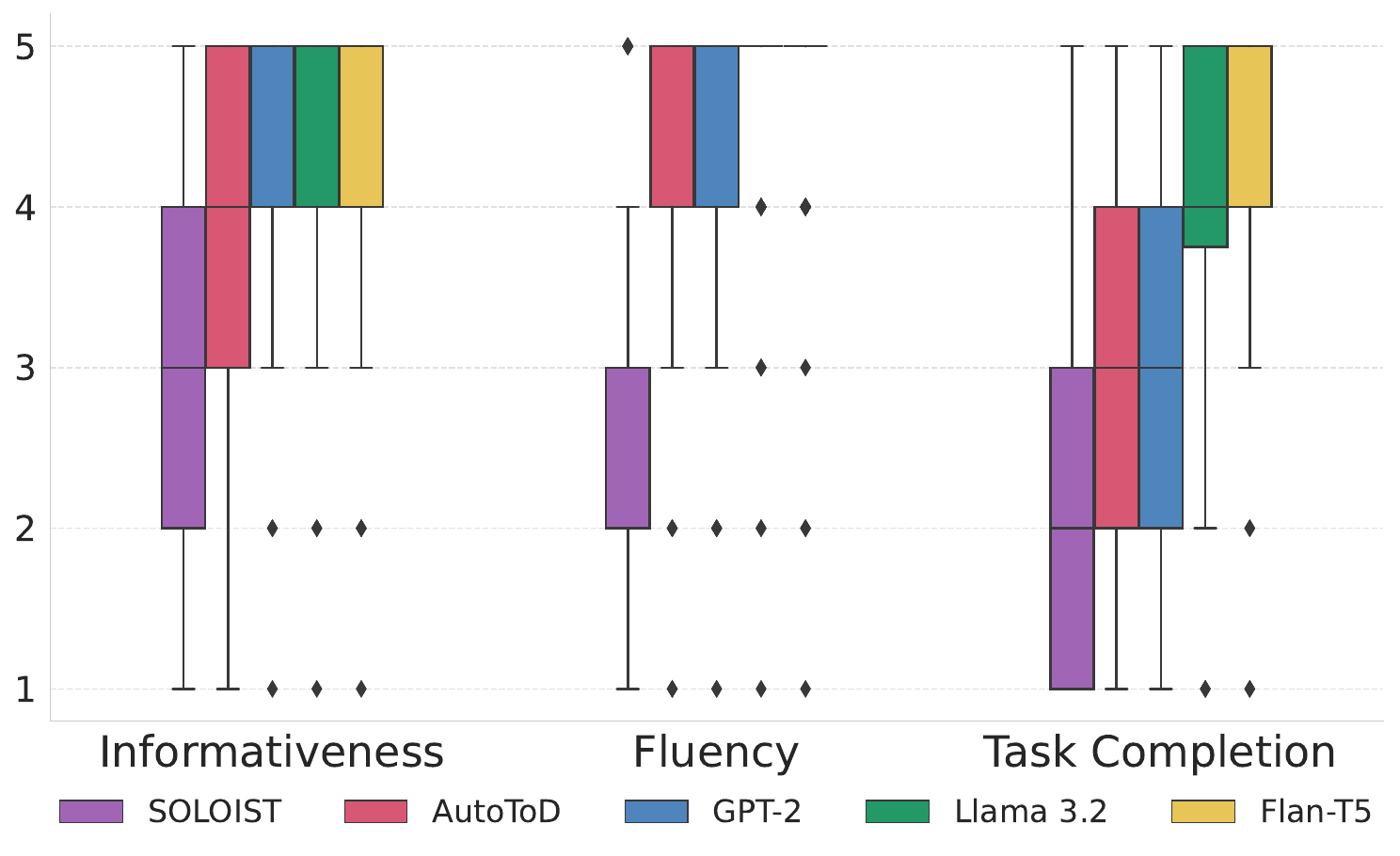}
   
   \caption{
       Human Evaluation Study on SGD and KETOD. Evaluators were asked to rate the dialog samples between a range of 1-5 on 3 categories.
   }
   \vspace{-15pt}
   \label{fig:human_evaluation}
\end{figure}
\vspace{-2pt}

\vspace{3pt}
\noindent
\textbf{Human Evaluation.} 
To supplement the automatic metrics and get a qualitative analysis, we conducted a human evaluation using Amazon Mechanical Turk to assess the performance of various models. Two baseline models ({\soloist} and {\autotod}) and three {\oursys} models ({\gpt}, {\llamai}, and {\flan}) were taken into account. We sampled 100 dialogs from each dataset, with 50 coming from single-domain tasks and the remaining 50 from multi-domain tasks, all from the test dataset. Human evaluators were asked to rate the models on a scale from 1 to 5 on three questions: the accuracy of information presented in the responses (Informativeness), how fluent and natural the conversation is (Fluency), and whether the models can make accurate {\apicall} (Task Completion).

The results, shown in Figure~\ref{fig:human_evaluation}, align with the automatic metrics, where {\oursys} models outperform the existing SOTA approaches. This demonstrates a strong alignment between quantitative and qualitative assessments. Notably, for task completion and fluency, {\llamai} and {\flan} demonstrate superior performance compared to all other models, which is consistent with our previous findings. Another important observation is that {\llamai} and {\flan} have less variance in performance across all tasks when compared to all other models, which further solidifies the robustness of our approach. 

% \fix{Should we not have more to say in human study? we do not even discuss each metric individually.
% which dataset these results belong to?}

\vspace{-6pt}
\section{Conclusion}
\vspace{-4pt}

\vspace{-3pt}
\begin{table}[!t]
    \centering
    \begin{adjustbox}{max width=\columnwidth}
    \small
        \begin{tabular}{|c|c|c|c|c|}
            \hline
            \makecell{\textbf{Dataset}} & \makecell{\textbf{Model}} & \makecell{\textbf{Overall} \\ \textbf{Response}} & \makecell{\textbf{API} \\ \textbf{Invoke} \\ \textbf{Accuracy}} & \makecell{\textbf{Complete} \\ \textbf{API} \\ \textbf{Accuracy}} \\ \hline
            \multirow{4}{*}{SGD} 
            & \autotod & 0.5471 & 63.15 & 42.20 \\
            & \gpt  & 0.7068 & 91.26 & 33.75 \\
            & \llamai & \textbf{0.7506} & \textbf{98.80} & \underline{53.68} \\
            & \flan  & \underline{0.7196} & \underline{94.42} & \textbf{61.07} \\ \hline
            
            \multirow{4}{*}{KETOD} 
            & \autotod & 0.5471 & 63.22 & 41.61 \\
            & \gpt & 0.6372 & \underline{94.48} & 25.52 \\
            & \llamai & \textbf{0.6454} & \textbf{97.70} & \underline{48.05} \\
            & \flan & \underline{0.7050} & \underline{92.18} & \textbf{59.54} \\ \hline
        \end{tabular}
    \end{adjustbox}
    \vspace{-6pt}
    \caption{Evaluation of fine-tuned approaches against large-scale proprietary LLMs on unseen domains (RQ3).}
    \label{tab:autotod_results}
    \vspace{-14pt}
\end{table}

This work demonstrates that LLMs fine-tuned solely on natural language dialogs can effectively generalize to unseen domains by framing ToD as a multi-task instruction fine-tuning problem.
To further enhance their {\ood} task completion performance, we introduce schema augmentation, which improves model adaptability to unseen domains and strengthens task completion performance. To ensure robust evaluation of task completion, we explicitly incorporate {\apicall}s as a core task and assess performance using both automatic metrics and human evaluations.
Furthermore, we show that fine-tuned ToD systems generalize better to unseen domains than fine-tuning-free approaches that rely on large-scale proprietary LLMs.
These results highlight the feasibility of developing cost-effective, scalable, and zero-shot generalizable ToD systems that achieve strong {\ood} generalization without requiring turn-level annotations, paving the way for their practical adoption in real-world applications.

%Our results show that fine-tuning not only improves task completion accuracy but also enables smaller, domain-adapted models that outperform larger proprietary LLMs, offering a more effective and cost-efficient alternative for ToD applications.

% In this work, we explored how ToD systems can function without annotated data by framing ToD as a multi-task instruction learning problem. Through schema augmentation, we enhanced the {\ood} generalization of ToD systems, improving their adaptability to unseen domains.
% To ensure effective task completion, we incorporate {\apicall}s as a core task and evaluate performance using automatic metrics. We also compared against fine-tuning-free approaches, and demonstrated that fine-tuning significantly enhances performance in complex domains.
% Our approach presents a scalable and adaptable alternative to traditional ToD models, reducing dependency on manual annotations while maintaining strong generalization and robustness.

% \fix{BLEU scores in appendix, sample dialogs in appendix, full prompt in appendix}
\section{Limitations}

{\oursys} has been developed by fine-tuning LLMs such as {\gpt}, {\llamai}, and {\flan}. 
These LLMs require significant computational resource requirements to train, particularly {\llamai}. 
Training and inference with these models can be expensive, limiting their practicality for deployment in resource-constrained environments.

%We remove earlier parts of the conversation to fit the context size of the models, thus making handling long-term dependencies in long multi-turn conversations a challenge for the system. 
%This limitation is a significant challenge for deploying the system in real-world, multi-turn dialog applications.
The LLMs used in the system function as black boxes, making it challenging to interpret the reasoning behind their responses. 
This lack of transparency hinders the ability to diagnose and correct erroneous outputs, which is crucial in ToD systems where accuracy is critical. Furthermore, the models may inherit biases present in the training data, leading to biased or unfair responses in certain scenarios. 
Although efforts were made to mitigate this issue by fine-tuning using the dialog datasets, completely eliminating biases remains a challenging task. %particularly when utilizing pre-trained models. 
The reliance on pre-trained models introduces limitations related to the coverage of the pre-training data. If the pre-training data lacks specific domain knowledge, the ToD system may under perform in those domains.

The deployment of LLMs in ToD systems raises ethical and privacy concerns, particularly regarding the handling of sensitive user data. Ensuring that the system complies with privacy regulations and ethical standards is an ongoing challenge that requires continuous monitoring and updates. 
Similar to other AI technologies, there is a scope for potential misuse of our system. 
If {\oursys} is used with malicious intent or the model is fed inappropriate data, there is a risk of abuse. 
We would strongly advise to take necessary precautions and appropriate usage policies.

Addressing the limitations outlined above is crucial for advancing the effectiveness and reliability of ToD systems.
While the usage of pre-trained LLMs offers significant advantages, these models are not without their challenges. 
Increasing model interpretability, mitigating biases, and addressing ethical and societal concerns are essential steps toward creating more robust and responsible ToD systems. 
%By acknowledging these limitations, we set the stage for continued research and innovation in the development of TOD systems.

\bibliography{my_files/custom}

\onecolumn
\appendix

% \section{Example Appendix}
% \label{sec:appendix}

% This is an appendix.
\section{Template for multi-task instruction fine-tuning}
\label{sec:templates}

Figure~\ref{fig:finetuning_template} shows the template used to process the inputs for {\oursys}. The template first informs about the domains involved in the dialog. Next, it provides task-specific instructions and provides the domain schemas. The dialog history is provided and the model is instructed to generate a system response based on the dialog history, search results, and the task instructions.

% \begin{figure*}
%     \centering
%     \includegraphics[width=0.97\linewidth]{assets/finetuning_template.pdf}
%     \caption{
% Multi-task instruction finetuning template}
%     \label{fig:finetuning_template}
% \end{figure*}

\begin{figure}[htbp]
    \centering
\begin{tcolorbox}[colframe=gray!20, colback=gray!10, coltitle=black, arc=5mm, boxrule=0.4mm, width=\columnwidth, left=3mm, right=3mm, top=3mm, bottom=3mm]
\setlength{\baselineskip}{1.5em}
You are an expert chat assistant for the domains: \textcolor{cyan!90}{[domains]}. \\
\textcolor{purple!90}{Instructions:} As an expert, you must generate the most appropriate response for the chat assistant. \\
The response can be an api call or a response to the user. \\
Based on the \textcolor{purple!90}{Last User Utterance}, you must find the relevant \textcolor{purple!90}{Intent} from the \textcolor{purple!90}{Schema} and your request should use the \textcolor{purple!90}{required slots} and \textcolor{purple!90}{optional slots} from that \textcolor{purple!90}{Intent}. \\
You will be provided with the Schema for domains: \textcolor{cyan!90}{[domains]}\\
\textcolor{cyan!90}{[schemas]}\\
You will be provided an incomplete dialog between a user and a chat assistant, and an optional search results.\\
\textcolor{purple!90}{Dialog History}: \textcolor{cyan!90}{[dialog history]}\\
Using the \textcolor{purple!90}{Dialog History}, \textcolor{purple!90}{Search Results}, and by following the \textcolor{purple!90}{Instructions} please generate the response for the chat assistant.
\end{tcolorbox}
    \caption{Multi-task instruction finetuning template. Items in blue are dynamic elements and those in purple are important aspects of the prompt.}
    \label{fig:finetuning_template}
\end{figure}
\section{BLEU Scores for Response Generation}
\label{sec:appendix_bleu}

Table~\ref{tab:response_blue} presents additional metrics for response generation. {\bleu} scores are reported for baselines ({\simpletod}, {\soloist}, {\zstod}, {\autotod}), and {\oursys} models (\gpt, \llamai, \flan) with schema augmentations.
We see a similar trend here as well, with {\oursys} models outperforming baseline approaches. 
However, {\bleu} scores are better for {\llamai} than {\flan}, particularly for the seen domains. 
Since the {\bleu} metric is calculated by n-gram matches, {\llamai} having better supervised performance tends to generate responses closer to the ground truth, thus yielding higher {\bleu} scores.

\begin{table*}[!t]
    \centering
    \begin{adjustbox}{max width=\textwidth}
        \begin{tabular}{|c|c|cccc|cccc|cccc|}
            \hline
    \multirow{2}{*}{\textbf{Dataset}} & \multirow{2}{*}{\textbf{Model}} & \multicolumn{4}{c|}{\textbf{Overall Response (BLEU-4)}} & \multicolumn{4}{c|}{\textbf{Inform (BLEU-4)}} &  \multicolumn{4}{c|}{\textbf{Request (BLEU-4)}} \\ %\cline{4-11}
    & & \textbf{All} & \textbf{Seen} & \textbf{Mixed} & \textbf{Unseen} & \textbf{All} & \textbf{Seen} & \textbf{Mixed} & \textbf{Unseen} &\textbf{All} & \textbf{Seen} & \textbf{Mixed} & \textbf{Unseen} \\ \hline
\multirow{7}{*}{SGD}     & \simpletod                                  
                        &  0.1696       &  0.1834      &     0.1877    & 0.1494      
                        &   0.1685      & 0.1790       & 0.1896       & 0.1438       
                        & 0.0228        & 0.0195       & 0.0216       & 0.0243       \\
                         & \soloist                                    
                         & 0.1902        & \underline{0.2798}        &0.1990         & 0.1655        
                         & 0.1813       & 0.2226       & 0.1945       & 0.1568
                         & 0.0281       & 0.0339        & 0.0265       & 0.0284       \\
                         & \zstod  
                         & 0.0590        & 0.1413        & 0.0568         &  0.0512
                         & 0.0255       & 0.0402       & 0.0228       & 0.0246
                         & 0.0231        & 0.0367       & 0.0221       & 0.0215       \\
                         & \autotod & 0.0487  & 0.0523  & 0.0501  & 0.0466 & 0.0854 & 0.0743 & 0.0884 & 0.0851 & 0.0173 & 0.0111 & 0.0159 & 0.0195 \\
                         \cline{2-14}
                         & \gpt & 0.2015  & 0.2109  & \underline{0.2229}  & 0.1802 & 0.2181 & 0.2421 & 0.2368 & 0.1923 & 0.0400   & 0.0275 & 0.0423 & 0.0403 \\
                         & \llamai & \textbf{0.2445}  & \textbf{0.2905}  & \textbf{0.2568}  & \textbf{0.2242} & \textbf{0.2888} & \textbf{0.3180}  & \textbf{0.3043} & \textbf{0.2650}  & \textbf{0.0641} & \textbf{0.0803} & \textbf{0.0614} & \textbf{0.0634} \\
                         & \flan  & \underline{0.2110}   & 0.2332  & 0.2226  & \underline{0.1961} & \underline{0.2811} & \underline{0.3098} & \underline{0.2911} & \underline{0.2631} & \underline{0.0569} & \underline{0.0625} & \underline{0.0541} & \underline{0.0582} \\
                         \thickhline
\multirow{7}{*}{KETOD}   & \simpletod                                  
                        & 0.0821        & 0.1015        & 0.0910        & 0.0538
                        & 0.1147       & 0.1362  & 0.1268       &  0.0726
                        & 0.0178       & 0.0266       & 0.0149       & 0.0106       \\
                         & \soloist                                    
                         & 0.0970        & 0.1018        & 0.0945       & 0.0848
                         & 0.0957       & 0.1185       &   0.0933     & 0.0675   
                         & 0.0167       & 0.0145       &      0.0174 & 0.0185  \\
                         & \zstod                                      
                         & 0.0394        & 0.0439        &   0.0254      & 0.0385
                         & 0.0183       & 0.0231       & 0.0059       &  0.0250
                         & 0.0260       & 0.0328       & 0.0198       & 0.0243       \\
                         & \autotod                                    & 0.0480   & 0.0528  & 0.0492  & 0.0415 & 0.0797 & 0.0678 & 0.0932 & 0.0812 & 0.0134 & 0.0157 & 0.0151 & 0.0092 \\
                         \cline{2-14}
                         & \gpt                                       & 0.1890   & 0.2106  & 0.1961  & 0.1524 & 0.2105 & 0.2437 & 0.2078 & 0.1687 & 0.0346 & 0.0500   & 0.0252 & 0.0263 \\
                         & \llamai & \textbf{0.2398}  & \textbf{0.2864}  & \textbf{0.2354}  & \textbf{0.1862} & \textbf{0.2701} & \textbf{0.3165} & \underline{0.2579} & \underline{0.2208} & 0.0581 & \underline{0.0723} & \textbf{0.0508} & \textbf{0.0490}  \\
                         & \flan & \underline{0.2082}  & \underline{0.2351}  & \underline{0.2048}  & \underline{0.1792} & \underline{0.2727} & \underline{0.3025} & \textbf{0.2811} & \textbf{0.2234} & \underline{0.0526} & \textbf{0.0750}  & \underline{0.0454} & \underline{0.0339}
                         \\
                         \hline
        \end{tabular}
    \end{adjustbox}
    \vspace{-6pt}
    \caption{BLEU Scores for Overall Response Generation, Inform and Request.}
    \label{tab:response_blue}
    \vspace{-6pt}
\end{table*}

\section{Dialog Examples}
\label{sec:appendix_dialogs}

Table~\ref{tab:rest_dialog} shows an example dialog in the Restaurant domain. The table contains the turn id, user utterance, gold response, {\soloist}, {\autotod}, {\gpt}, {\llamai}, and {\flan} response. 
Text highlighted in red outlines the portions where the system response by a model is incorrect and green highlights the correct parts. Texts highlighted in orange indicate that the model is partially correct and is missing some information.

In the example dialog, we can see that the responses of {\autotod} is longer and more descriptive, whereas {\oursys} models produce succinct responses, similar to the gold standards. This observation indicates that {\oursys} was able to pick up the text generation style of the gold standard, which is an important aspect for ToD systems.

The baseline {\soloist} fails to generate a valid response on a few turns. This stems from the complex structure of the output, which consists of dialog states, actions and system response. The output structure has a certain format and different parts are identified with special tokens. On many occasions, the model does not produce the correct tokens, thus leading to invalid generations. 

On turns where the user asks about multiple information, the {\gpt} and {\soloist} model struggles to respond to all the user requests. However, all the other models do not have this problem. This could be attributed to the fact that {\gpt} models are the smallest model when compared to all the other models.

{\llamai} and {\flan} generates responses that are very similar to the gold standard. These models are able to pickup the generation style and can respond to multiple requests of the user. In the example dialog, the user provided information about the restaurant name, time and date on turn 1, but then changed the restaurant name on turn 2. {\llamai} and {\flan} models were able to identify what changed and moved forward with the {\apicall} by using all the correct information. However, {\autotod} was not able to keep track of all the changes and made an incorrect {\apicall}. {\soloist} was able to make a partially correct {\apicall} since it was not able to generalize to the \texttt{number\_of\_seats} slots. It used the \texttt{party\_size} slot that it had seen during training. 

This dialog example showed some of the drawbacks of baseline approaches and some example cases of where they struggle. For {\oursys} models, the limitations of {\gpt} was evident, as it was not able to process multiple requests of the user. However, {\llamai} and {\flan} had performance similar to the gold standard across multiple turns, which falls in line to our automatic metrics and human evaluation.

% \small
\begin{longtable}{>{\raggedright}p{2cm}|p{13cm}}
% \begin{longtable}{r|l}
    \caption{Dialog Example---SGD dataset. Dialog id 1\_00001} \label{tab:rest_dialog}\\  
    \hline
    \textbf{Model} & \textbf{Response---SGD dataset. Dialog id 1\_00001}  \\  
    \hline
    \endfirsthead

    % Header on subsequent pages
    \hline
    \textbf{Dataset} & \textbf{Response---SGD dataset. Dialog id 1\_00001 }  \\  
    \hline
    \endhead

    % Footer for all pages except the last
    \hline
    \textbf{Continued on next page...} & \\  
    \hline
    \endfoot

    % Footer for the last page
    \hline
    \endlastfoot

    % Table rows with long text
Turn & 1 \\ \hline
            User & Can you book a table for me at the Ancient Szechuan for the 11th of this month at 11:30 am?  \\ 
            Gold & In which city are you trying to book the table? \\ 
            \soloist & \mycolor{BrickRed}{Please confirm the following details: Booking a table for 2 at Ancient Szechuan in San Francisco on March 11th at 11:30 am.} \\ 
            \autotod & \mycolor{ForestGreen}{Could you please provide the city where the restaurant `Ancient Szechuan'' is located and the number of seats you would like to reserve?}
 \\
            \gpt & \mycolor{BrickRed}{Sure, I have a table for 2 at Ancient Szechuan in San Francisco at 11:30 am today.}
 \\ 
            \llamai & \mycolor{ForestGreen}{What is the location of the restaurant?} \\
            \flan &  \mycolor{ForestGreen}{Where should I look for the restaurant?}
\\ 
      \hline      
            Turn & 2 \\ \hline
            User &  Can you book a table at the Butterfly restaurant in San Francisco?  \\ 
            Gold & Please confirm that you need a table for 2 at the Butterfly Restaurant in San Francisco at 11:30 am on March 11th. \\ 
            \soloist & Can you please confirm that you want to book a \mycolor{ForestGreen}{table for 2 at the Butterfly restaurant in San Francisco at 11:30 am on March 11th?}\\ 
            \autotod &  \mycolor{BurntOrange}{Could you please confirm the date and time for the reservation at the Butterfly restaurant in San Francisco? Additionally, let me know the number of seats you would like to reserve.}
\\
            \gpt & \mycolor{BrickRed}{Can you please confirm that you want to book a table for 2 people at the Butterfly restaurant in San Francisco at 11:30 am on March 11th?}
 \\ 
            \llamai & Can you confirm the following details: You want to book a \mycolor{ForestGreen}{table at the butterfly Hotel in San Francisco at 11:30 am. You want 2 seats. You want to book the table on March 11th.} \\

            \flan & You wish to book a \mycolor{ForestGreen}{table for 2 at the Butterfly Restaurant in San Francisco for March 11th at 11:30 am.} Is this correct?
 \\ 
            \hline
            Turn & 3 \\ \hline
            Gold & ApiCall(method=`ReserveRestaurant', parameters= `date': `2019-03-11', `location': `San Francisco', `number\_of\_seats': `2',`restaurant\_name': `Butterfly Restaurant', `time': `11:30' )\\ 
            \soloist & ApiCall(method='ReserveRestaurant', parameters={`city': 'San Francisco', `date': `2019-03-11', \mycolor{BurntOrange}{`party\_size'}: `2',`restaurant\_name': `The Butterfly Restaurant', `time': `11:30'}) \\ 
            \autotod & ApiCall(method=`FindRestaurants',parameters=\mycolor{BrickRed}{`category': `Butterfly'}, `location': `San Francisco')
 \\
            \gpt & ApiCall(method=`ReserveRestaurant', parameters=`date': `2019-03-11', `location': `San Francisco', \mycolor{ForestGreen}{`number\_of\_seats'}: `2',`restaurant\_name': `The Butterfly Restaurant', `time': `11:30') \\
            \llamai & ApiCall(method=`ReserveRestaurant', parameters=`date': `2019-03-11', `location': `San Francisco', \mycolor{ForestGreen}{`number\_of\_seats'}: `2',`restaurant\_name': `Butterfly Restaurant', 'time': `11:30')
\\ 
            \flan & ApiCall(method=`ReserveRestaurant', parameters= `date': `2019-03-11', `location': `San Francisco',`restaurant\_name': `Butterfly Restaurant', \mycolor{ForestGreen}{`number\_of\_seats'}: `2', `time': `11:30' )
 \\ 
\hline            
            Turn & 4 \\ \hline
            User & That's perfect. What cuisine do they offer?  \\ 
            Gold & Your table has been booked successfully and they serve Asian cuisine. \\ 
            \soloist & \mycolor{BurntOrange}{Your table has been booked.} \\ 
            \autotod & It looks like the reservation at the Butterfly Restaurant in San Francisco has been successfully made for March 11th at 11:30 am for 2 people. \mycolor{ForestGreen}{The restaurant offers Asian cuisine.} If you need any more assistance or have any other questions, feel free to ask. Have a great day! Goodbye!
  \\
            \gpt & \mycolor{BurntOrange}{Your table is booked.}
\\
\llamai & Your reservation has been successful. \mycolor{ForestGreen}{They offer Asian cuisine.}
\\
            \flan & Your reservation has been made. \mycolor{ForestGreen}{They serve Asian food.} \\ 

            \hline
            Turn & 5 \\ \hline
            User & Can you get me the user rating of the restaurant? Is it expensive? \\ 
            Gold & Their prices are moderate and they have a user rating of 4.0. \\ 
            \soloist & \failed \\ 
            \autotod & The Butterfly Restaurant in San Francisco has a user \mycolor{ForestGreen}{rating of 4.00. The price range is moderate}, so it's not considered expensive. If you have any more questions or need further assistance, feel free to ask!
\\
            \gpt & \mycolor{Orange}{The rating is 4.0.}
 \\
 \llamai & The restaurant has a \mycolor{ForestGreen}{4.0 rating and is considered moderate in price}.
\\
            \flan &  They have a \mycolor{ForestGreen}{moderate price range and a 4.0 rating}.
\\

            Turn & 6 \\ 
            User & Thanks a bunch! \\ 
            Gold & Is there anything else I can do for you? \\ 
            \soloist & \failed \\ 
            \autotod & You're welcome! If you need any more assistance, feel free to ask. Have a great day! Goodbye!
 \\
            \gpt & Is there anything else I can help you with?\\
            \llamai & Do you need anything else? \\
            \flan & Do you need help with anything else?
 \\

            Turn & 7 \\ 
            User & No. Thank you. That's all I need. \\ 
            Gold & Have a great day! \\ 
            \soloist & Have a great time at the restaurant! \\ 
            \autotod & You're welcome! If you need any more assistance in the future, feel free to reach out. Have a great day! Goodbye!
\\
            \gpt & Have a good one!
\\
\llamai & Have a great day. \\
            \flan & Have a nice day.
 \\ 

\end{longtable}

% \begin{longtable}{>{\raggedright}p{2cm}|p{13cm}}
% % \begin{longtable}{r|l}
%     \caption{Dialog Example---SGD dataset. Dialog id 1\_00001} \\  
%     \hline
%     \textbf{Model} & \textbf{Response---SGD dataset. Dialog id 1\_00001}  \\  
%     \hline
%     \endfirsthead

%     % Header on subsequent pages
%     \hline
%     \textbf{Dataset} & \textbf{Response---SGD dataset. Dialog id 1\_00001 }  \\  
%     \hline
%     \endhead

%     % Footer for all pages except the last
%     \hline
%     \textbf{Continued on next page...} & \\  
%     \hline
%     \endfoot

%     % Footer for the last page
%     \hline
%     \endlastfoot
    
%     % Table content using tabular
%     \begin{tabular}{r|p{13cm}}

%     \end{tabular}

%     \end{longtable}
\section{User Study Instructions}
\label{sec:templates}

\section*{Disclaimers of any risks to participants or annotators}

There are no significant risks associated with participating in this study. However, annotators may experience mild fatigue or cognitive strain due to prolonged reading and evaluation of multiple conversations. If you feel discomfort or fatigue, please take breaks as needed.

\section*{Instructions for Human Study Participants}

Your task is to evaluate model-generated responses in multi-turn, task-oriented conversations based on the following criteria:
\begin{enumerate}
    \item Fluency
    \item Informativeness
    \item Task Completion
\end{enumerate}

\section*{Task Overview}
\begin{itemize}
    \item You will be presented with multiple conversations, where a user interacts with a model to complete a task (e.g., booking a flight).
    \item Your job is to rate each model response independently using a 1-5 scale based on the provided criteria.
    \item This study is blind --- you will not know which model produced which response.
\end{itemize}

\section*{How to Rate Responses}
You will assign a score for each response based on the following:

\subsection*{1. Fluency}
Measures whether the response is grammatically correct, well-structured, and natural to read.

\begin{itemize}
    \item 1 = Unnatural, grammatically incorrect, or hard to read.
    \item 5 = Perfectly fluent, natural, and error-free.
\end{itemize}

\textbf{Example (Good Fluency)}
\begin{quote}
User: I need to book a flight to New York next Monday.\\
Model: Sure! What time would you like to depart?
\end{quote}

\textbf{Example (Poor Fluency)}
\begin{quote}
User: I need to book a flight to New York next Monday.\\
Model: Sure! You next Monday flight want time what?
\end{quote}

\subsection*{2. Informativeness}
Measures whether the response provides useful and relevant information to advance the task.

\begin{itemize}
    \item 1 = Vague, incorrect, or unhelpful.
    \item 5 = Precise, relevant, and useful.
\end{itemize}

\textbf{Example (Good Informativeness)}
\begin{quote}
User: Can you recommend a vegetarian restaurant nearby?\\
Model: Yes! Green Leaf Café is a highly rated vegetarian restaurant.
\end{quote}

\textbf{Example (Low Informativeness)}
\begin{quote}
User: Can you recommend a vegetarian restaurant nearby?\\
Model: There are many restaurants in your area.
\end{quote}

\subsection*{3. Task Completion}
Measures whether the response successfully progresses or completes the task in the conversation.

\begin{itemize}
    \item 1 = Fails to address the request.
    \item 5 = Fully completes the task.
\end{itemize}

\textbf{Example (Good Task Completion)}
\begin{quote}
User: I’d like to book a table for two at an Italian restaurant at 7 PM today.\\
Model: I found a reservation at Bella Italia for 7 PM. Would you like me to reserve it for 2 people at 7 PM today?
\end{quote}

\textbf{Example (Failed Task Completion)}
\begin{quote}
User: I’d like to book a table for two at an Italian restaurant at 7 PM.\\
Model: There are Italian restaurants in your area.
\end{quote}

\section*{Additional Guidelines}
\begin{itemize}
    \item Be objective --- rate based on quality, not personal preference.
    \item If a response is unclear or ambiguous, leave a comment.
    \item Do not try to guess which model produced the response.
\end{itemize}

\section*{Study Duration \& Submission}
\begin{itemize}
    \item The study will take approximately 10 minutes to complete.
    \item Once you have evaluated all responses, submit your ratings.
\end{itemize}

\vspace{1cm}
\centering{\textbf{Thank you for your time and valuable feedback!}}

\end{document}